\documentclass[conference]{IEEEtran}

\usepackage{graphicx} 
\usepackage{amsmath} 
\usepackage{cite}
\usepackage{lipsum}
\usepackage[font=small,skip=2pt]{caption}

\begin{document}

\title{Stepping Stabilization Using a Combination of DCM Tracking and Step Adjustment}

\author{\IEEEauthorblockN{Majid Khadiv\IEEEauthorrefmark{1}\IEEEauthorrefmark{2}\IEEEauthorrefmark{4},
Sebastien Kleff\IEEEauthorrefmark{2},
Alexander Herzog\IEEEauthorrefmark{2},\\
S. Ali. A. Moosavian\IEEEauthorrefmark{1},
Stefan Schaal\IEEEauthorrefmark{2}\IEEEauthorrefmark{3}, and
Ludovic Righetti\IEEEauthorrefmark{2}}
\IEEEauthorblockA{\IEEEauthorrefmark{1}Department of Mechanical Engineering, K. N. Toosi University of Technology, Tehran, Iran}
\IEEEauthorblockA{\IEEEauthorrefmark{2}Autonomous Motion Department, Max-Planck Institute for Intelligent Systems, Germany}
\IEEEauthorblockA{\IEEEauthorrefmark{3}CLMC Lab, University of Southern California, Los Angeles, USA}\\
\IEEEauthorblockA{\IEEEauthorrefmark{4}Corresponding author: mkhadiv@mail.kntu.ac.ir, majid.khadiv@tuebingen.mpg.de}}

\maketitle

\begin{abstract}
In this paper, a method for stabilizing biped robots stepping by a combination of Divergent Component of Motion (DCM) tracking and step adjustment is proposed. In this method, the DCM trajectory is generated, consistent with the predefined footprints. Furthermore, a swing foot trajectory modification strategy is proposed to adapt the landing point, using DCM measurement. In order to apply the generated trajectories to the full robot, a Hierarchical Inverse Dynamics (HID) is employed. The HID enables us to use different combinations of the DCM tracking and step adjustment for stabilizing different biped robots. Simulation experiments on two scenarios for two different simulated robots, one with active ankles and the other with passive ankles, are carried out. Simulation results demonstrate the effectiveness of the proposed method for robots with both active and passive ankles.
\end{abstract}


\section{Introduction}
In order to be able to take part in our future daily life, humanoid robots should be capable of safely performing various tasks in highly dynamic environments. This capability requires a real-time walking pattern generation unit to produce feasible walking patterns in complicated environments. Furthermore, the generated patterns should be robust against uncertainties and disturbances. 

Exploiting the whole dynamics of a biped robot for generating walking patterns demands high computation burden, while convergence to the global minimum is not guaranteed\cite{herzog2015trajectory,khadiv2015optimal}. However, simple models may be employed to generate walking patterns in real-time. The Linear Inverted Pendulum Model (LIPM) \cite{kajita20013d}, reduces the Center of Mass (CoM) dynamics of a biped robot to a linear model by assuming a constant CoM height and negligible angular momentum. This model has been very successfully used for the design of walking controllers for complex biped robots. Kajita et al.  \cite{kajita2003biped} proposed a preview control of the Zero Moment Point (ZMP) to generate a CoM trajectory based on a predefined ZMP trajectory. Wieber\cite{wieber2006trajectory} improved the performance of this approach in the presence of relatively severe pushes. He proposed to recompute the trajectories in a Model Predictive Control (MPC) framework without any predefined ZMP trajectory, using feedback from the current state of the robot. Furthermore, other formulations of the walking pattern generation as an MPC problem have been suggested, for example by considering step locations in the optimization procedure \cite{herdt2010online}. 

Analytical methods have been presented which do not rely on optimization algorithms \cite{morisawa2006biped,harada2006analytical}. In these methods, position and velocity of the CoM are considered as states of the system. Then, the trajectory for the CoM is obtained, using the analytical solution of the LIPM and continuity constraints at the control points. Although considering the position and velocity of the CoM as states of the system enables us to generate feasible walking patterns, the problem is over-constrained. The reason is that by constraining the position and velocity, both divergent and convergent parts of the LIPM dynamics are constrained. Hence, Takaneka et al. \cite{takenaka2009real} divided the LIPM dynamics  into its divergent and convergent parts. Then, they just constrained the divergent part to generate the Divergent Component of Motion (DCM) trajectory based on predefined footprints (ZMP trajectory). Before \cite{takenaka2009real}, the divergent part of the LIPM dynamics had been used by Hof et al.\cite{hof2008extrapolated} to explain human walking properties under the name of extrapolated Center of Mass (XCoM). This concept was also developed in Pratt et al. \cite{pratt2006capture} under the name of Capture Point (CP), the point on which the robot should step to come to rest. 

By employing the DCM, Englsberger et. al \cite{englsberger2015three} proposed a method to control the unstable part of the CoM dynamics without affecting the stable part. Although this controller can react to the disturbances very fast, perfect DCM tracking needs unconstrained manipulation of the CoP. For robots with finite size feet and actuated ankles, the CoP can be directly manipulated inside the support polygon, using the ankle actuation. However, since the size of the feet is limited, the actual control authority of the ankle joint is also limited. This problem is more severe, when the robot has point contact feet or finite size feet with passive ankles. For a robot with point contact feet, the CoP is always located at the point of contact during stepping. As a result, manipulation of the CoP is not possible. For robots with passive ankles, the CoP cannot be manipulated directly by the ankle joint torques and has to be manipulated by more proximal joints which renders its control more difficult.

 In addition to CoP manipulation, step adjustments constitute a significant tool for stabilizing biped robots and these adjustments can be realized in a slower time scale (e.g. every step). A constraint of step adjustment is that the step location is limited by the reachable area, especially in very constrained environments.

In order to map control policies generated with the simplified LIPM dynamics to the full robot, a whole body controller is employed. For robots with torque-control capability, methods based on inverse dynamics can be used \cite {mistry2010inverse,righetti2013optimal}. In this context, Herzog et al. \cite{herzog2016momentum} proposed a method for solving the Hierarchical Inverse Dynamics (HID) and demonstrated its performance through various balance experiments on a torque controlled biped robot. The HID enables us to take into account equality and inequality constraints, and also to prioritize tasks with respect to each other. The latter property is especially useful for our work to control biped robots with different structures.

In this work, we propose an approach for stabilizing biped robots stepping by a combination of DCM tracking and step adjustment strategies (Fig. 1). We employ different combinations of the DCM tracking and step adjustment in the HID, to stabilize biped robots with active and passive ankles. Furthermore, for step adjustment, a swing foot trajectory modification method is proposed. The rest of this paper is organized as follows: in section II the desired trajectory for the DCM is generated. Section III proposes a method for swing foot trajectory modification. Section IV introduces the HID and the considered tasks. In section V the simulation results are presented and discussed. Finally, section VI concludes the findings.

\section{Trajectory Generation for the DCM}
\begin{figure}
\centering
\includegraphics[clip,trim=7cm 19.35cm 7cm 2.4cm,width=8cm]{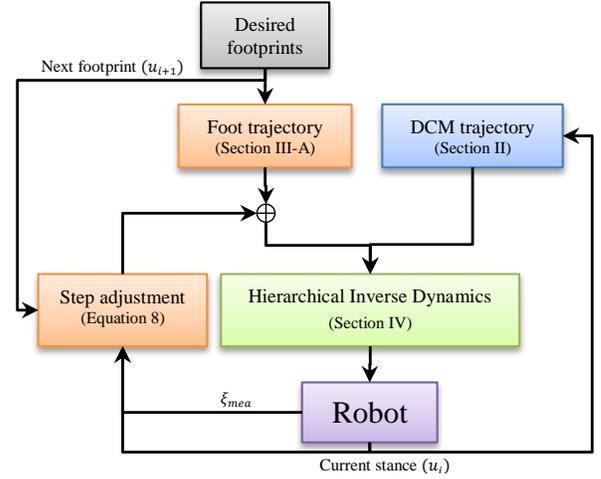}
\caption{The proposed control framework}
\vspace{-1.5em}
\label{block_diagram}
\end{figure}

The block diagram of the proposed walking controller is shown in ‎Fig. \ref{block_diagram}. As it may be observed, both desired feet and DCM trajectories are generated. Based on feedback from the current state of the robot, the foot trajectory is modified to compensate for the DCM tracking error. The generated trajectories, then, are fed into the HID to generate desired actuating torques of the joints consistent with the specified tasks and hierarchy. 

The LIPM constrains motion of the CoM on a plane (horizontal plane for walking on a flat surface), by using a telescopic massless link connecting the CoP to the CoM \cite{kajita20013d}. The dynamics of this system may be formulated as:
\begin{equation}
\label{LIPM}
\ddot{x} = \omega_0^2 (x-u)
\end{equation}
in which $x$ is a 2-D vector containing CoM horizontal components (the vertical component has a fixed value $ z_0$), and $u$ is the CoP vector ($u=[CoP_x,CoP_y]^T$).  Furthermore, $\omega_0$ is the natural frequency of the pendulum ($\omega_0=\sqrt[]{g/z_0}$, where $g$ is the gravity constant, and $z_0$ is the CoM height).

By considering CoM ($x$) and DCM ($\xi=x+\dot{x}/\omega_0$) as the state variables, the LIPM dynamics in the state space form may be specified as:
\begin{align}
\label{DCM_CoM}
\begin{split}
&\dot{x} = \omega_0 (\xi-x) \\ 
&\dot{\xi} = \omega_0 (\xi-u)
\end{split}
\end{align}

Equation (\ref{DCM_CoM}) decomposes the LIPM dynamics into its stable and unstable parts, where the CoM converges to the DCM and the DCM is pushed away by the CoP. Hence, in order to have a stable walking pattern, it is enough to constrain the DCM motion during walking, without restricting the other state of the system.

Constraining the DCM motion can be achieved in different ways, i.e. applying the DCM boundary condition to have cyclic motion \cite{takenaka2009real}, specifying the DCM at the end of a previewed number of steps to be at the top of foot print \cite{englsberger2015three}, and etc. In this paper, we use the method in \cite{englsberger2015three}, and set the DCM (its projection on the ground) coincident with the footprint at the end of a previewed number of steps.

By solving the second equation of (\ref{DCM_CoM}), the DCM trajectory based on the natural dynamics of the LIPM can be obtained:
\begin{equation}
\label{initial_value}
\xi(t) = (\xi_0-u) e^{\omega_0 t}+u
\end{equation}

where $\xi_0$ and $u$ are the DCM and CoP at the beginning of a step. In the last step in the previewed period, we force the DCM to be coincident with the last footprint. Then, we compute the required initial condition for the DCM:
\begin{equation}
\label{Boundary_DCM}
\xi_{0,n-1} = (u_n-u_{n-1}) e^{-\omega_0 T}+u_{n-1}
\end{equation}

where $n$ is the number of previewed steps, and $T$ is the step period. Then, as it is illustrated in ‎Fig. 2, we compute the DCM boundary conditions in a recursive fashion: 
\begin{equation}
\label{recursive_DCM}
\xi_{0,i-1} = (\xi_{0,i}-u_{i-1}) e^{-\omega_0 T}+u_{i-1} \quad , \quad  i=1,...,n-1.
\end{equation}

Using this method, DCM boundary conditions which constrain the DCM motion are computed. It should be mentioned that the CoP is considered fixed during each step. Furthermore, since this recursive procedure is carried out at each step, the next footprints can be changed during the previewed period. In other word, the boundary conditions are updated at each step, based on the updated footprints.

\section{Step Adjustment }

The desired DCM trajectory should be tracked to realize the generated walking pattern. In other word, deviations from the desired DCM trajectory may cause instability. However, robots with different structures have different authorities to manipulate the CoP. For example, a robot with passive ankles has to use joints above the knee in order to modulate the CoP, whereas robots with actuated ankles can apply ground forces directly at the foot. As a result, relying on the DCM tracking is not enough for stepping stabilization of robots with different structures in the presence of disturbances and uncertainties. In this section, we propose a method for swing foot trajectory modification, by using the DCM feedback. The aim of this real-time swing foot trajectory modification is to realize the adapted landing point at the end of the step.

\subsection{Swing Foot Trajectory }
The trajectory for the swing foot is generated to ensure that the foot lands without impact or slip. For the horizontal components of the swing foot, we consider fifth order polynomial to satisfy position, velocity and acceleration constraints at the start and end of the swing phase. For the vertical direction, beside these constraints, the maximum height of the swing foot at the midpoint of the swing phase is specified. As a result, a sixth order polynomial is employed to satisfy all the constraints in the vertical direction. These trajectories are computed at the start of each step for a previewed period. 

\subsection{Real-time Modification}
\begin{figure}
\centering
\includegraphics[clip,trim=7cm 19.5cm 7cm 2cm,width=7cm]{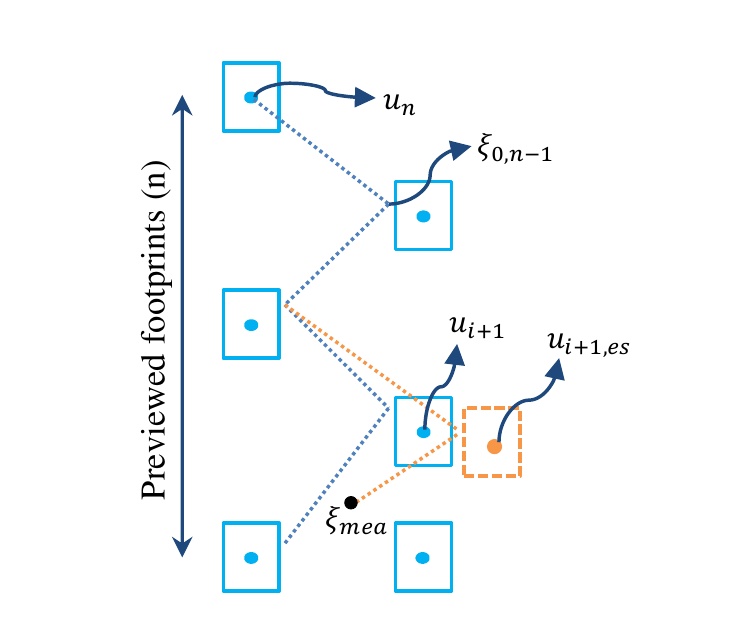}
\caption{the DCM boundary condtition and landing point adjustment}
\vspace{-1.5em}
\label{step_adjustment}
\end{figure}

‎Figure \ref{step_adjustment} illustrates the adaptation procedure that we propose for the swing foot trajectory. In this procedure, the goal is to modify the swing foot landing in the current step (using measured DCM) such that the new footprint brings the DCM to the desired DCM boundary condition at the end of the next step (Fig. \ref{step_adjustment}). Using this method, if there are no more disturbances, after correction the swing foot will land on the predefined footprint, in the next steps.

By rearranging (\ref{initial_value}), and setting the current DCM measurement as the initial state, the DCM position at the end of the current step can be obtained as follows:
\begin{equation}
\label{DCM_prediction}
\xi_{0,i+1,es} = (\xi_{mea}-u_{i}) e^{\omega_0 (t-T)}+u_{i} \quad , \quad  0 \leq t \leq T 
\end{equation}

in which $\xi_{mea}$ is the measured DCM, and $\xi_{0,i+1,es}$ is the estimated DCM at the end of the current step. It is important to note that in the case of perfect DCM tracking, $\xi_{0,i+1,es}$ would coincide with $\xi_{0,i+1}$ computed from the recursive equation of (\ref{recursive_DCM}).

Now, we aim at finding the next footprint which brings the DCM from $\xi_{0,i+1,es}$ to $\xi_{0,i+2}$ , for the next step. Again, rearranging (\ref{initial_value}) and solving it for the desired next footprint yields:
\begin{equation}
\label{next_step}
u_{i+1,es} =\frac{ \xi_{0,i+2}-\xi_{0,i+1,es} \: e^{\omega_0 T}}{1- e^{\omega_0 T}} \quad , \quad  0 \leq t \leq T 
\end{equation}
where $u_{i+1,es}$ is the estimated next footprint for compensating the tracking error of the DCM and bringing it to its predefined boundary condition in the next step.

Now, the swing foot modification can be computed:
\begin{equation}
\label{modification}
mod=u_{i+1,es}-u_{i+1}
\end{equation}
This modification is added to the swing foot trajectory in the horizontal directions, at each control cycle. In our step adjustment method, the trajectory of the swing foot in the vertical direction is not modified. As a result, at the start of each step, the trajectories for the feet are generated for the previewed period. Then, the modification from (\ref{modification}) is computed in real-time based on DCM measurement, and added to the horizontal components of the swing foot trajectory. 

\section{WHOLE BODY CONTROL}
In order to obtain feasible joint torque commands from the generated trajectories in the last sections,we use a hierarchical inverse dynamics controller. In HID framework, constraints and tasks are written as affine functions of joint and floating-base accelerations, actuation torques, and interaction forces and moments. In fact, these variables constitute the design variables which are optimized through a series of Quadratic Programs (QP), at each control cycle. The goal of the controller is to find the design variables that satisfy different control objectives with different priorities, while the highest priority in the hierarchy is set to ensure physical consistency. In lower priorities, various tasks with different ranks are specified, and tasks in the same priority can be weighted with respect to each other \cite{herzog2016momentum}.

For all the tasks, we put the equations of motion and the limits for the actuation in the highest priority. The equations of motion can be written as:
\begin{equation}
\label{dynamics_eq}
M(q) \ddot{q}+N(q,\dot{q})=S^T \tau+J_c^T \lambda
\end{equation}

in which $M$ is the inertia matrix, $q$ is the vector of generalized coordinates, and $N$ groups together the Coriolis, centrifugal and gravitational effects. $S$ represents the joint selection matrix, $\tau$ is the vector of actuation torques, $J_c$ is the contact  jacobian, and $\lambda$ is the vector of contact forces and moments. In the following, we describe lower priority tasks and constraints that are meant to track swing foot and DCM profiles, using admissible contact forces.
\subsection{Foot Trajectory Tracking}
The modified trajectory for the swing foot in section III should be tracked to realize the stepping task. This task can be written as: 
\begin{equation}
\label{swing_control}
J_{sw} \ddot{q}+\dot{J}_{sw} \dot{q}=\ddot{X}+K_d(\dot{X}_d-\dot{X})+K_p(X_d-X)
\end{equation}
in which $X$ and $X_d$ are the actual and reference swing foot posture vectors , and $J_{sw}$ is the jacobian of the swing foot. Furthermore, $K_p$ and $K_d$ are diagonal gain matrices. It should be noted that the desired orientation components of the swing foot is set to zero. For the stance foot, the task is:
\begin{equation}
\label{stance_control}
J_{st} \ddot{q}+\dot{J}_{st} \dot{q}=0
\end{equation}

where $J_{st}$ is the jacobian of the stance foot. This task keeps the stance foot in a stationary contact with the ground surface.

\subsection{DCM Tracking}
For the DCM tracking task, we employ the following control rule \cite{englsberger2015three}:

\begin{equation}
\label{DCM_control}
\dot{\xi}-\dot{\xi}_d=-K_{\xi}(\xi-\xi_d)
\end{equation}

in which $k_{\xi}$ is the DCM control gain, while $\xi$ and $\xi_d$ are the measured and desired DCM, respectively. Substituting this equation into the second equation of (\ref{DCM_CoM}) yields:

\begin{equation}
\label{desired_CoP}
u_{des}=\xi+\frac{1}{\omega_0}(K_{\xi}(\xi-\xi_d)-\dot{\xi}_d)
\end{equation}

where $u_{des}$ is the desired CoP location that ensures the closed-loop behavior defined in (\ref{DCM_control}). It should be noted that the desired CoP computed from (\ref{desired_CoP}) is not necessarily feasible, and should be projected inside the support polygon (CoP constraint, subsection C). This projection results in a DCM tracking error that is compensated using step adjustment (section III).

The obtained desired CoP from (\ref{desired_CoP}), can be related as a CoM task using (\ref{LIPM}):
\begin{equation}
\label{CoM_control}
\ddot{X}_{CoM, ref}= \begin{bmatrix} \omega_0^2(x-u_{des})\\0 \end{bmatrix}=J_{CoM} \ddot{q}+\dot{J}_{CoM} \dot{q}
\end{equation}
where $J_{CoM}$ specifies the CoM jacobian, and $x$ is a 2-D vector containing CoM horizontal components.

\subsection{Contact Constraints}
The CoP can be computed from the contact forces ($\lambda$). To guarantee stationary contact, the CoP should lie inside the support polygon, which introduces an inequality constraint. Furthermore, the resultant interacting forces at the contact points should stay inside the friction cones. The friction cones are approximated by pyramids to yield linear inequality constraints on the contact forces \cite{herzog2016momentum}.
\subsection{Posture Control}
Posture control is a task in the joint space. We use this task to constrain the robot to be as much as possible in upright posture during walking. This task can be written as:

\begin{equation}
\label{joint_control}
\ddot{q}=K_p(q_d-q)-K_d \dot{q}
\end{equation}

In this equation, $K_p$ and $K_d$ are diagonal square gain matrices, with the same dimension as the actuated joints.
\begin{figure}
\centering
\includegraphics[clip,trim=7cm 20cm 7cm 2.6cm,width=8cm]{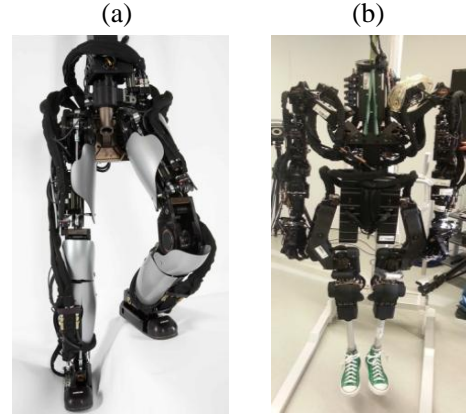}
\caption{The two Sarcos humanoid robots used in simulation (a) Hermes, which has fully actuated ankles (Credit:Luke Fisher Photography) (b) Athena, which has passive ankles }
\vspace{-1.5em}
\label{robot}
\end{figure}
\subsection{Base Control}
The base control task keeps the torso in an upright position. It is defined as::
\begin{equation}
\label{base_control}
J_{base} \ddot{q}+\dot{J}_{base} \dot{q}=K_p(Q_d-Q)-K_d \dot{Q}
\end{equation}

where $J_{base}$ is the rotational part of the floating base jacobian. Furthermore, $Q$ and $Q_d$ are the actual and desired quaternion of the floating base.

\section{Results and Discussions}
In this section, we present two scenarios in a simulation environment to show the effectiveness of the proposed framework. In these scenarios, we employ different hierarchies in the HID consistent with the control authority in the ankles.The first scenario corresponds to implementing walking patterns on a fully actuated biped robot (‎Fig. \ref{robot}(a)). In this scenario, we push the robot to show the robustness of the gaits in the presence of disturbances. In the second scenario, we remove the actuation of the ankles (‎Fig. \ref{robot} (b)), and obtain a stable stepping in place. Then, we change the time of stepping and demonstrate the effectiveness of step adjustment to stabilize the robot motion.
\begin{figure*}
\centering
\includegraphics[clip,trim=2.5cm 16.4cm 2cm 2cm,width=13.5cm]{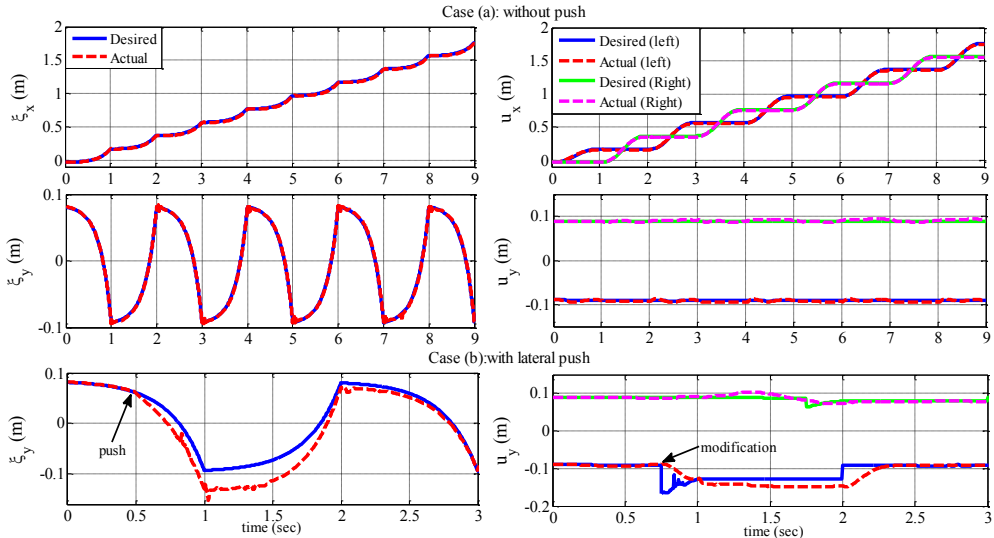}
\caption{First scenario: walking of Hermes with active ankles on a flat surface (in simulation environment). Case (a): The sagittal and lateral components of the DCM and feet without disturbance are shown. The desired DCM is tracked without need for step adjustment. Case (b): the lateral components of the DCM and feet in the presence of a push are shown (the push force is 10 percent of the robot weight and its duration is 0.3 second). The step adjustment could deal with the disturbance and stabilize the walking. }
\vspace{-1.5em}
\label{active}
\end{figure*}

\subsection{Walking with Actuated Ankles}
In the first scenario, simulation of walking with the lower part of Sarcos humanoid robot Hermes (‎Fig. \ref{robot}(a)) is performed. Since the robot is equipped with active ankle joints, we put the DCM tracking task in the same rank as the foot trajectory tracking task, in the first rank in the hierarchy (physical consistency constraints are at rank 0). Besides, in order to constrain the controller to exploit feasible forces for the DCM tracking, we put the contact constraints (CoP and friction constraints) in the first rank, as well. The posture control and base control tasks are put in the lower priority to exploit the remaining degrees of freedom to keep the robot in an upright posture as much as possible.

\begin{figure*}
\centering
\includegraphics[clip,trim=2.5cm 16.8cm 2cm 2cm,width=14cm]{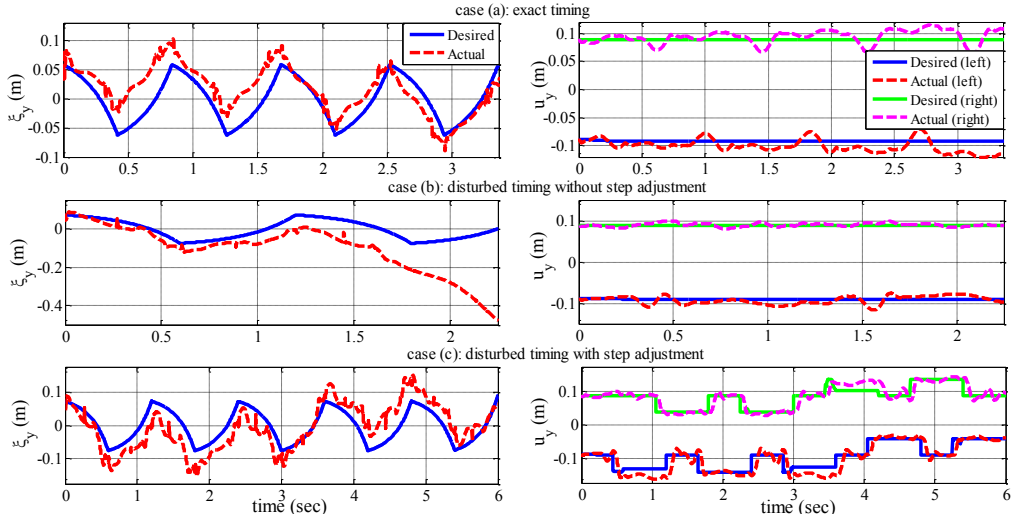}
\caption{Second scenario: Stepping in place of Athena with passive ankles (in simulation environment).In this scenario the lateral components of the  DCM and feet for stepping in place are shown. Case (a): the step timing is tuned to yield a stable stepping without controlling the DCM. Case (b): the tuned step timing is disturbed by 50 percent, so the DCM trajectory diverges and the robot is unstable. Case (c): Step adjustment is activated and stabilizes the robot for the disturbed case.}
\vspace{-1.5em}
\label{passive}
\end{figure*}
In ‎Fig. \ref{active} case (a), the desired and actual values of the DCM and feet trajectories in both sagittal and lateral directions are depicted. As it can be observed, in this case, since the DCM trajectory is tracked, modification of the swing foot trajectory is not required. Hence, the desired trajectories of the feet remain almost constant, during normal walking. In order to show the robustness during walking, in the case (b), the robot is pushed laterally (the push force is 10 percent of the robot weight and its duration is 0.3 second). As it can be observed in the left figure, the actual DCM diverges from the desired trajectory. The reason is that in this case the forces required for tracking the DCM are infeasible and projecting them into feasible area causes tracking error. In order to compensate for this tracking error, the swing foot trajectory is modified (the right figure), using the algorithm described in section III. The result of this modification can be observed as the DCM converges to its desired value in the next step. 
\subsection{Stepping in place with passive ankles}
In the second scenario, we simulate stepping in place of the Sarcos humanoid robot Athena (‎Fig. \ref{passive}(b)), with passive ankles. In this scenario, having the DCM tracking task in the highest priority would be problematic since the controller would create large arms and upper-body motions to change the CoP. It is not a desired behavior. Furthermore, there is no constraint for step adjustment, because the environment is not constrained. As a result, we consider another hierarchy consistent with the actuation in the ankles. We put the foot trajectory tracking and the posture control in the highest priority. Because we put the posture control in the highest priority, there is no more degrees of freedom to control the DCM. As a result, in this case, we rely just on step adjustment to stabilize the stepping.

The obtained results from this scenario are illustrated in ‎Fig. \ref{passive}. We only present trajectories in the lateral direction since the robot is stepping in place. For the case (a), the step timing is selected such that following the natural dynamics of the robot realizes the stepping. As it can be seen in the case (a), the foot trajectory in the lateral direction is not modified. However, the measured DCM does not diverge, and remains on a limit cycle. Although the stepping in this case is stable, this situation is fragile; because a slight change in the step timing causes instability of the robot. To show this, we increase the step timing by 50 percent in the second case. The obtained results are shown in ‎Fig. \ref{passive}(b). As it can be observed, without step adjustment , the DCM trajectory diverges. This divergence causes instability of the robot.

In order to show the robustness of the stepping exploiting step adjustment as a stabilizing tool, we repeat the case (b) with step adjustment. As a result, the landing point of the swing foot is modified to realize stable stepping. By comparing the DCM trajectory in the cases (b) and (c), it is clear that the DCM does not diverge in the case (c). In fact, by modifying the swing foot trajectory and changing the landing position consequently, the DCM trajectory remains on a limit cycle. This scenario shows significance of the proposed step adjustment method for preserving stability of the robot. 

\section{CONCLUSIONS}
In this paper, a method for stabilizing biped robots stepping based on a combination of DCM tracking and step adjustment was proposed. This method is based on generating a DCM trajectory consistent with footprints, and adjusting step locations using DCM feedback. For step adjustment, a novel strategy based on modifying the swing foot trajectory was suggested. To use a combination of these two stabilizing tools consistent with the robot control authority, a hierarchical inverse dynamics controller was used. Two simulation scenarios were performed to demonstrate the effectiveness of the proposed method. In the first scenario, a robot with active ankles was able to stabilize walking using a combination of DCM tracking and step adjustment. In the second scenario, a robot with passive ankles stabilized stepping by using only step adjustments without any explicit CoM or DCM tracking task. The obtained results showed that the proposed method can stabilize stepping of robots with passive and active ankles, in the presence of disturbances.

\section*{Acknowledgment}
This research was supported by the Max-Planck Society, MPI-ETH center for learning systems and the European Research Council under the European Union's Horizon 2020 research and innovation program (grant agreement No 637935).

\bibliography{Master}
\bibliographystyle{IEEEtr}

\end{document}